%% file: main.tex
\documentclass[conference]{IEEEtran}

\input epsf
\usepackage{graphicx}
\usepackage[sorting=none]{biblatex}
\usepackage{amsmath,amssymb}
\usepackage{algorithm}
\usepackage{algorithmicx}
\usepackage{algpseudocode}
\usepackage{booktabs} 
\usepackage{verbatim}

\begin{document}

\title{\LARGE FedReplay: A Feature Replay Assisted Federated Transfer Learning Framework for Efficient and Privacy-Preserving Smart Agriculture }

\author{\IEEEauthorblockN{Long Li\IEEEauthorrefmark{1},Jiajia Li\IEEEauthorrefmark{2}, Dong Chen\IEEEauthorrefmark{3}, Lina Pu\IEEEauthorrefmark{4}, Haibo Yao\IEEEauthorrefmark{5}, *Yanbo Huang\IEEEauthorrefmark{5} }
\\
\IEEEauthorblockA{\IEEEauthorrefmark{1} Department of Electrical and Computer Engineering, The University of Alabama \\
                    \IEEEauthorrefmark{2} Electrical and Computer Engineering, Michigan State University \\
                     \IEEEauthorrefmark{3} Agricultural and Biological Engineering, Mississippi State University \\
                     \IEEEauthorrefmark{4} Department of Computer Science, University of Alabama \\
                     \IEEEauthorrefmark{5} USDA-ARS Genetics and Sustainbale Agriculture \\
}
}

\maketitle
\renewcommand{\thefootnote}{} 
\footnotetext{* Corresponding author : yanbo.huang@usda.gov}
\addtocounter{footnote}{-1}   

\begin{abstract}
Accurate classification plays a pivotal role in smart agriculture, enabling applications such as crop monitoring, fruit recognition, and pest detection. However, conventional centralized training often requires large-scale data collection, which raises privacy concerns, while standard federated learning struggles with non-independent and identically distributed (non-IID) data and incurs high communication costs. To address these challenges, we propose a federated learning framework that integrates a frozen Contrastive Language–Image Pre-training (CLIP) vision transformer (ViT) with a lightweight transformer classifier. By leveraging the strong feature extraction capability of the pre-trained CLIP ViT, the framework avoids training large-scale models from scratch and restricts federated updates to a compact classifier, thereby reducing transmission overhead significantly. Furthermore, to mitigate performance degradation caused by non-IID data distribution, a small subset (1\%) of CLIP-extracted feature representations from all classes is shared across clients. These shared features are non-reversible to raw images, ensuring privacy preservation while aligning class representation across participants. Experimental results on agricultural classification tasks show that the proposed method achieve 86.6\% accuracy, which is more than 4 times higher compared to baseline federated learning approaches. This demonstrates the effectiveness and efficiency of combining vision-language model features with federated learning for privacy-preserving and scalable agricultural intelligence.
\end{abstract}
\IEEEoverridecommandlockouts

\IEEEpeerreviewmaketitle

\input{Sections/introduction.tex}

\input{Sections/background.tex}

\input{Sections/preliminary.tex}

\input{Sections/method.tex}

\input{Sections/result.tex}

\input{Sections/conclusion.tex}

\printbibliography

\end{document}

%% file: Sections/introduction.tex
\section{Introduction}
Smart agriculture has emerged as a transformative approach to addressing the challenges of sustainable food production, resource efficiency, and labor reduction \cite{vzalik2023review,esfandiari2025multi}. With the integration of advanced sensing and computing technologies, smart agriculture enables applications such as crop growth monitoring, fruit detection and classification, weed and pest identification, and yield estimation \cite{li2024secure, ahmed2024smart, li2024soybeannet}. These tasks often rely on accurate classification and object detection models \cite{li2022reinforcement,yuan2025smart, li2021q}. By leveraging artificial intelligence (AI), particularly computer vision and machine learning techniques, agricultural systems can achieve enhanced productivity, timely disease control, and optimized resource allocation, thus contributing to food security and environmental sustainability \cite{herterich2025accelerating}.

Despite these advantages, training effective machine learning models typically requires centralized access to large-scale datasets \cite{zhu2022gearbox}. Centralized training, however, often raises significant privacy concerns, as it requires the collection and storage of raw data, such as farm images or operational logs, at a central server \cite{li2025extra}. In many cases, agricultural data sharing is restricted by policies, regulations, or farmers’ reluctance due to ownership and privacy issues \cite{gavai2025agricultural}. To mitigate these challenges, federated learning (FL) has been proposed as a promising paradigm that enables multiple participants to collaboratively learn a global model without sharing raw data, thereby ensuring data privacy while promoting collaborative intelligence \cite{kondaveeti2024federated}.
\begin{figure}[htp]
    \centering
    \includegraphics[width=0.8\linewidth]{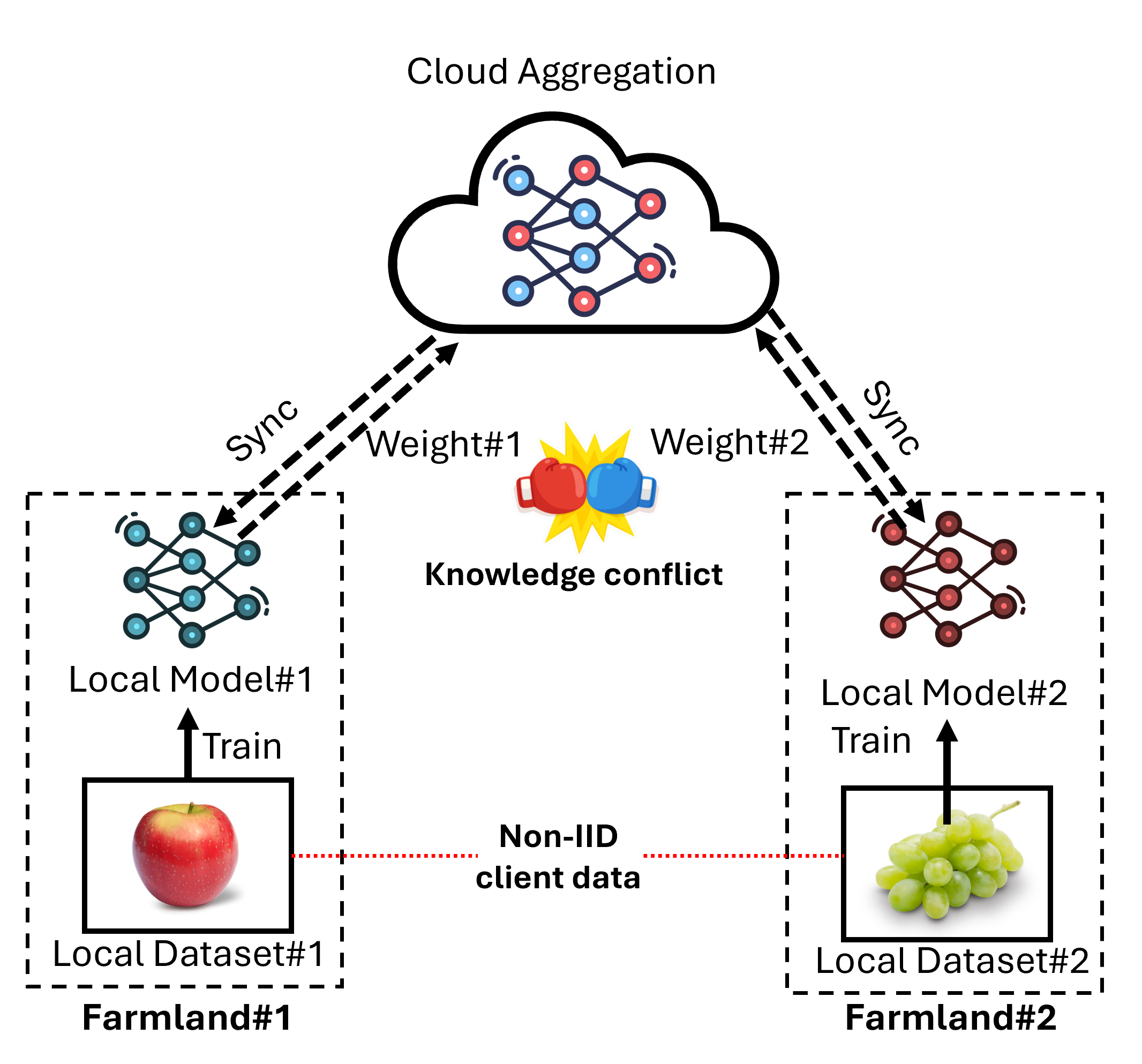}
    \caption{Illustration of knowledge conflict cause by non-IID dataset distribution in an FL framework}
    \label{fig:example}
\end{figure}

Nevertheless, conventional FL continues to confront several fundamental challenges. Fig. \ref{fig:example} presents a representative FL application framework within the agricultural domain.
The first issue limitation lies in the substantial communication overhead, which results from the iterative exchange of large-scale model parameters between distributed clients and the central server during the training process \cite{li2025vllfl}. This not only consumes substantial bandwidth but also leads to long training times, especially when deep neural networks are trained from scratch, as shown in Fig. \ref{fig:example}. The second issue arises from the non-independent and identically distributed (non-IID) nature of client data \cite{dembani2025agricultural}. In agriculture, the heterogeneity of environments, crop types, and imaging conditions results in imbalanced and divergent local datasets \cite{vzalik2023review}. Training on non-IID data often slows convergence and may degrade the performance of the global model, as conflicting gradient updates across clients compromise the overall learning process \cite{lu2024federated}, as illustrated in Fig. \ref{fig:example}.

To address these limitations, prior research has explored multiple strategies. Communication-efficient FL approaches reduce overhead by compressing gradients, quantizing parameters, or implement transfer learning technique \cite{gugssa2025enabling,wang2022communication}. Other works have attempted to alleviate the non-IID issue by employing data augmentation, knowledge distillation, or client clustering strategies to align distributions across participants \cite{zhang2021fedpd}. While these methods provide partial relief, they either compromise model performance, require complex coordination, or fail to fundamentally reduce the training burden of deep models. Hence, a novel framework that simultaneously addresses both the communication efficiency and the non-IID distribution problem is highly desirable.

In this study, we propose a new framework, feature replay assisted federated transfer learning framework (FedReplay), which leverages transfer learning with a pre-trained vision-language model (VLM) to enhance federated learning for agricultural classification. Instead of training deep models from scratch, our framework employs a frozen CLIP vision transformer (ViT) to extract robust feature representations with rich semantic knowledge learned from large-scale datasets. A lightweight transformer classifier is then trained on top of these frozen features in the FL setting, greatly reducing the size of model updates exchanged across clients. To further mitigate the non-IID problem, FedReplay shares a very small subset (1\%) of feature embeddings, rather than raw images, across clients. These shared features act as reference points that align class representations globally, while ensuring privacy preservation since they are non-reversible to original images.

Extensive experiments on agricultural classification benchmarks demonstrate that FedReplay achieves 86.6\% accuracy, representing more than a four-fold improvement compared with standard FL baselines. Moreover, the framework reduces communication overhead by approximately 98\% relative to conventional FL approaches, highlighting its efficiency and scalability. These results confirm the effectiveness of combining vision-language pre-training with federated learning for privacy-preserving, communication-efficient, and high-performance smart agriculture.
The contribution of this paper are as below:

\begin{enumerate}
    \item We propose a novel federated learning framework that integrates a pretrained vision–language model (CLIP) with a lightweight Transformer-based classifier through transfer learning. The framework enhances visual classification performance while significantly reducing communication overhead, enabling scalable and privacy-preserving deployment in smart agriculture applications.
    
    \item We design several strategies to mitigate data heterogeneity across clients, including a feature replay mechanism and a row-gated aggregation policy. A systematic procedure is also developed for seamless late-joining client integration without retraining the entire federation.
    \item We conduct extensive experiments to validate the effectiveness and efficiency of the proposed framework under realistic agricultural scenarios. The results demonstrate that our method achieves 86.6\% classification accuracy, representing over a 4 times improvement compared with baseline approach, while maintaining low communication cost and strong convergence stability. 
\end{enumerate}

This paper is organized as follow: Section. \ref{sec:background} provide a background introduction of related topic. Section. \ref{sec:preliminary} provide a preliminary test result to show the impact of Non-IID issue and discussion of feature replay. Section. \ref{sec:methods} provide detailed design of the proposed FL framework. In Section. \ref{sec:result}, we conduct a comprehensive evaluation and present the result. Section. \ref{sec:conclusion} concludes the paper.

%% file: Sections/background.tex
\section{Background}
\label{sec:background}
This section reviews the key technical foundations of our work. We introduce vision–language models, principles of federated learning and the challenges it faces, and discuss transfer learning as an effective strategy to adapt pretrained models for agricultural-specific applications while reducing training and communication costs.

\subsection{Vision Language Model}
Vision-Language Models (VLMs) represent a significant breakthrough in artificial intelligence, bridging the gap between visual perception and natural language comprehension \cite{li2023knowledge}. These models are pre-trained on vast internet-scale datasets of image-text pairs, allowing them to learn a shared semantic space where visual concepts are aligned with textual descriptions. The primary advantage of this approach is its powerful zero-shot generalization capability, which allows VLMs to understand and execute tasks on new, unseen data categories without any task-specific training or fine-tuning \cite{zhang2024vision}. This ability to perform open-vocabulary recognition has dramatically accelerated the development of versatile AI applications. In recent years, the field has evolved rapidly, moving from basic image-text matching to more sophisticated models capable of complex visual reasoning, dialogue, and instruction following \cite{li2024vision}.

The landscape of VLMs includes several foundational and state-of-the-art models that have driven progress in the field. CLIP was a pioneering model that demonstrated the effectiveness of contrastive learning for aligning images and text, setting a new standard for zero-shot image classification \cite{radford2021learning}. Building on this, subsequent models have integrated vision encoders with powerful Large Language Models (LLMs) to enhance their reasoning and interactive capabilities. For example, LLaVA (Large Language and Vision Assistant) combines the CLIP vision encoder with the Vicuna LLM, enabling it to engage in complex visual dialogue and follow instructions related to an image \cite{liu2023visual}. More specialized models like Grounding DINO have pushed the boundaries of open-set object detection, allowing users to locate any object in an image using free-form text queries \cite{liu2024grounding}.

The powerful capabilities of VLMs are increasingly being adapted for specialized domains, most notably in smart agriculture. Several recent studies have demonstrated their potential to revolutionize agricultural practices. In the work \cite{li2025metafruit}, they leverages a comprehensive multi-fruit dataset to benchmark and advance the development of foundation models specifically for agriculture. In \cite{zhang2025clip}, the authors propose E-CLIP, an enhanced CLIP-based model designed to achieve more accurate and robust fruit detection and recognition in complex orchard environments. To address crop health, a visual large language model was developed for in-the-wild wheat disease diagnosis \cite{zhang2024visual}, enabling farmers to identify issues directly from images. Furthermore, FSVLM \cite{10851315} presents a vision-language model tailored for segmenting farmland from remote sensing imagery, showcasing the utility of VLMs in large-scale land management.

Despite their promise, integrating VLMs into real-world agricultural applications faces significant challenges. A major drawback of the current paradigm is the necessity of centralized data collection, training or fine-tuning these models often requires amassing huge datasets from various farms, which is logistically challenging and introduces serious privacy concerns \cite{li2025benchmark}. Agricultural data is often proprietary and sensitive, and its sharing may be restricted by data privacy regulations or competitive business interests \cite{gavai2025agricultural}. This creates a critical need for methods that can leverage data from multiple sources without compromising privacy. To address this gap, our proposed federated learning method is built upon the CLIP model. We use CLIP as an example to demonstrate our approach because it is a foundational, powerful, and widely-understood VLM whose robust vision encoder serves as an excellent and representative baseline for developing and evaluating new techniques.

\subsection{Federated learning}
\label{subsec:background}
Federated Learning (FL) is a decentralized machine learning paradigm that enables collaborative model training without requiring access to raw, private data \cite{zhao2018federated}. Unlike conventional centralized training, where data from various sources is aggregated in a single server for processing, FL operates by keeping data localized on client devices \cite{wen2023survey}. The training process involves distributing a global model to multiple clients, training it locally on their private data, and then sending only the resulting model updates—such as weights or gradients—back to a central server for aggregation. The primary advantage of this approach is its inherent privacy preservation, making it ideal for applications where data is sensitive, proprietary, or subject to strict regulations \cite{wen2023survey}. Consequently, FL is particularly suitable for industries like healthcare, finance, and mobile services, where collaboration is beneficial but data sharing is not feasible.

\subsubsection{Federated Learning in Smart Agriculture}
The principles of Federated Learning are especially appropriate for agricultural applications, where farm data is often geographically dispersed, proprietary, and highly sensitive. This alignment has spurred significant research into applying FL to solve various agricultural challenges. For instance, RuralAI \cite{devaraj2024ruralai} introduced a hierarchical FL system for real-time tomato crop health monitoring. Researchers have also applied FL with Convolutional Neural Networks (CNNs) for soybean leaf disease detection \cite{rajput2024smart} and for assessing wheat disease severity \cite{mehta2023transforming}, enabling multiple farms to build robust models without sharing private imagery. Beyond crop health, FL is being used to create sustainable solutions, such as the FLAG framework for optimizing irrigation practices \cite{bera2024flag}. Furthermore, FL has been employed for agricultural risk management through collaborative crop yield prediction \cite{manoj2022federated} and to enhance security in agricultural IoT networks with systems like FELIDS \cite{friha2022felids}.

More recently, research has begun to explore the powerful combination of Federated Learning and Vision-Language Models (VLMs) to bring advanced multimodal understanding to decentralized environments. The VLLFL framework \cite{li2025vllfl}, for example, proposes a lightweight federated system using a VLM specifically for smart agriculture, aiming to balance performance with efficiency. To address the challenge of training large models, methods like FLoRA \cite{nguyen2024flora} have been developed to enhance VLMs using parameter-efficient federated learning techniques, which drastically reduce the amount of data that needs to be communicated. In the context of large-scale environmental monitoring, FedRSCLIP \cite{lin2025fedrsclip} demonstrates a federated approach for remote sensing scene classification, effectively leveraging the zero-shot capabilities of VLMs in a privacy-preserving manner.

\subsubsection{Challenge for Federatea Learning}
Despite its advantages, general Federated Learning suffers from critical issues that can hinder its practical deployment. First is the heavy communication overhead, the repeated exchange of model parameters between the server and numerous clients can be incredibly resource-intensive \cite{li2025vllfl}. This problem is magnified when using VLMs, whose parameter counts can number in the hundreds of millions, making the transmission of full model updates prohibitively expensive. Second, FL systems often experience a significant performance loss due to non-IID data \cite{lu2024federated}. In any real-world federated application, the data distribution on each client will inevitably be unique, which causes client models to diverge and their updates to conflict during server-side aggregation, ultimately degrading the global model's accuracy \cite{vzalik2023review}.

To address these challenges, existing research has proposed various solutions. To reduce communication overhead, techniques like gradient compression have been introduced, which apply methods like sparsification and quantization to shrink the size of model updates before transmission \cite{LinHM0D18},\cite{reisizadeh20a}. Another approach involves optimizing the training process itself; for instance, the original FedAvg algorithm proposed only updating and averaging a fraction of clients in each round to reduce the total communication load \cite{mcmahan2017communication}. To mitigate the non-IID issue, researchers have explored personalized federated learning, where techniques like model-agnostic meta-learning are used to train a global model that can be rapidly adapted to each client's local data distribution \cite{fallah2020personalized}. Another effective strategy is federated transfer learning, which leverages knowledge from pre-trained models to create more robust and generalizable client models, as demonstrated in the FedHealth framework for wearable healthcare data \cite{chen2020fedhealth}.

However, there is currently no single, unified solution that effectively resolves both communication overhead and the non-IID data problem, particularly for the enormous scale of modern VLMs. Conventional model pruning and compression techniques are often insufficient to make the federated training of massive VLMs practical. As a motivation to solve these challenges, especially in FL applications built upon VLMs, our proposed method takes both communication overhead and the non-IID issue into consideration. It is designed to solve these problems in tandem and boost performance, thereby benefiting the capability of advanced AI in agricultural applications.

\subsection{Transfer Learning}
Transfer learning enables a model to reuse knowledge acquired during large-scale pretraining and adapt it to a target task with modest data and compute \cite{chen2025towards}.  Its primary benefit is a significant reduction in training time and the ability to achieve high performance even with limited task-specific data \cite{gugssa2023enhancing}. In vision–language models (VLMs) such as CLIP, encoders trained on massive image–text pairs produce rich, transferable visual representations \cite{radford2021learning}. Freezing the CLIP vision encoder and fine-tuning a small classifer or detection head to exploits this knowledge without incurring the cost and instability of full-model retraining.

A growing literature combines transfer learning with FL to reduce communication overhead, especially for deploying large models like VLM. For example, the FedHealth framework demonstrated how leveraging knowledge from pre-trained models in an FL setting could build effective personalized healthcare models on wearable device data \cite{chen2020fedhealth}. In the context of VLMs, where full model fine-tuning is often infeasible due to communication constraints, parameter-efficient transfer learning methods are critical. Techniques like prompt tuning \cite{zhao2023fedprompt} or federated text-driven prompt generation \cite{qiu2024federated} allow clients to collaboratively learn optimal text prompts to steer the VLM's behavior without altering its core parameters. Similarly, adding small, trainable adapter modules \cite{nguyen2024flora} allows for model specialization by only training and transmitting a tiny fraction of the total parameters, making federated training of these massive models practical.

Motivated by these findings and the need to curb communication costs, this study adopts a transfer-learning design that freezes a pretrained VLM (CLIP ViT) encoder and trains only a compact classifier within the FL loop for agricultural tasks. Restricting updates to the small classifier reduces the number of trainable and transmitted parameters, which in turn shortens each communication round and lowers total rounds to reach target accuracy due to faster, more stable convergence on pretrained features. This architecture aligns with privacy and scalability requirements in smart agriculture while directly addressing both per-round overhead and end-to-end training time.

%% file: Sections/preliminary.tex
\section{Preliminary result and finding}
\label{sec:preliminary}

In this section, we will demonstrate a preliminary test to confirm the existence and evaluate the impact of Non-IID issue in traditional federated learning. And we will discuss the potential solution that can be applied to metigate the Non-IID issue.
\subsection{Non-IID Challenge in Federated Learning}

\label{sec:non-iid}
A fundamental challenge in FL is non-IID issue as discussed in Section. \ref{subsec:background}. In real-world agricultural applications, different clients may collect data from distinct environments, crop types, or imaging conditions, leading to heterogeneous and imbalanced local datasets. This discrepancy often results in slower convergence, unstable optimization, and reduced final accuracy.

\begin{figure}[htp]
    \centering
    \includegraphics[width=0.9\linewidth]{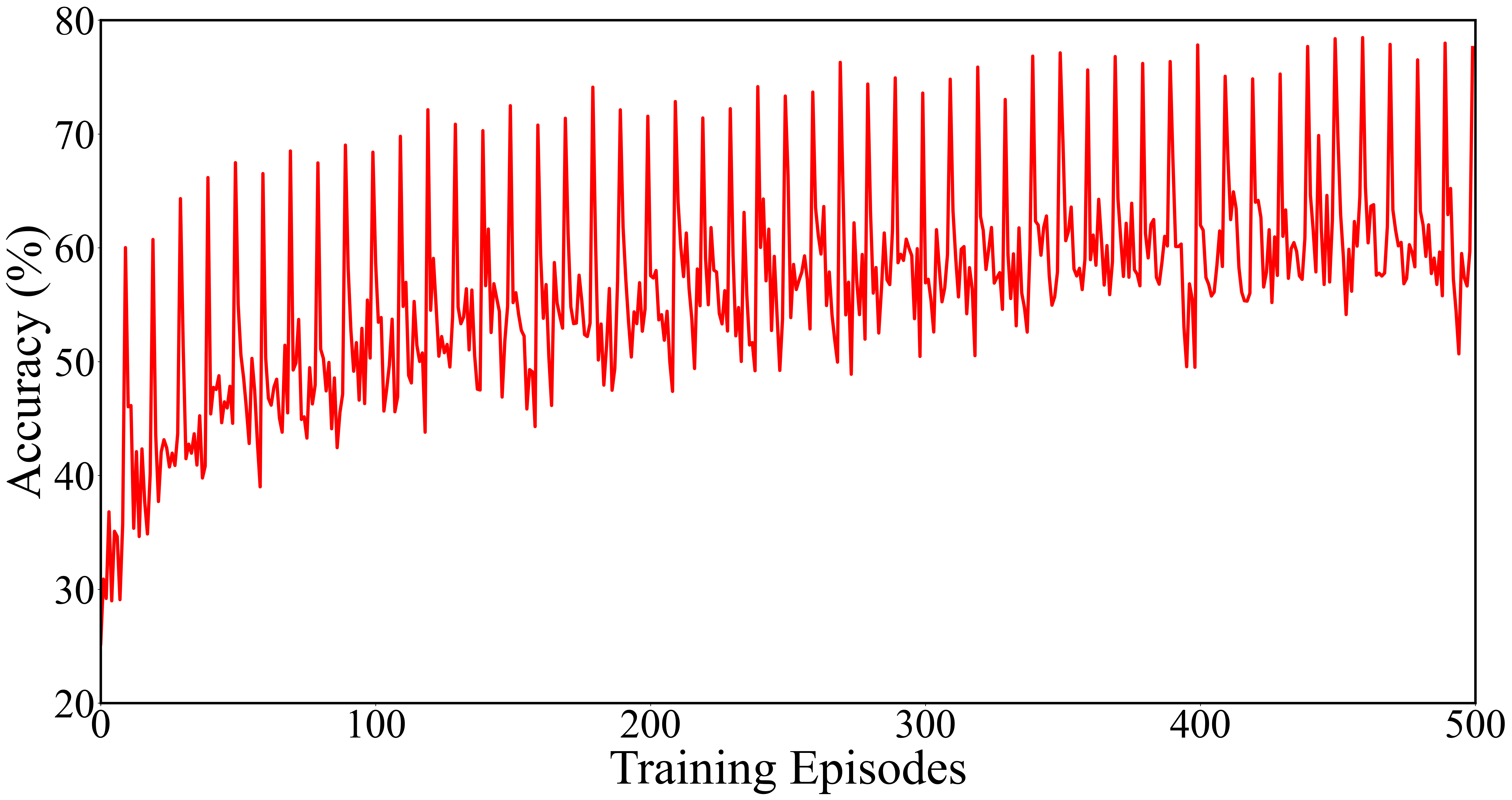}
    \caption{Illustration of how non-IID data distribution degrades federated training performance.}
    \label{fig:non-iid}
\end{figure}

To illustrate this effect, we conducted an experiment to demonstrate how non-IID issue impact the performance of FL. In this experiment, each client owned a unique subset of data, containing classes not present in other clients. In parallel, we assumed the server maintained access to the full dataset. We train the transformer classifier in conventional federated learning setting on a test dataset. But for every ten communication episodes, the global model was fine-tuned on the full dataset and redistributed to all clients. In theory, the global fine-tune should provide optimal update and boost the performance and federated learning performance should continuously improve on top of that if knowledge aggregation is also optimal. However, the result in Fig. \ref{fig:non-iid} tells a different story. Each pulse in the accuracy curve corresponds to the boost gained from global fine-tuning, which indicates the potential performance achievable with balanced data. However, subsequent federated aggregation consistently led to sharp drops in accuracy, demonstrating that conflicting updates from non-IID clients directly counteract the optimal gradient direction provided by the full-dataset fine-tuning. This validates that non-IID distributions significantly degrade the performance of FL and prevent the global model from sustaining the accuracy achieved during fine-tuning.

The detrimental effect of non-IID data has also been highlighted in prior studies. For example, \cite{zhao2018federated} demonstrated that sharing just 5\% of the global dataset among all clients can improve accuracy by 30\% on CIFAR-10, confirming the severity of performance degradation caused by non-IID data distribution. However, Sharing raw data across clients, even in small proportions, compromises the privacy guarantees that FL aims to uphold. 

To address this issue, our work introduces feature replay to mitigate the non-IID problem while maintain the privacy-preserving advantage of FL. Instead of sharing raw images, each client contributes a small subset of non-reversible features extracted by a frozen VLM encoder. These features provide a shared global reference that aligns class-level representations across clients, reducing the divergence between local updates. In this way, our method achieves the accuracy benefits of data sharing while maintaining the privacy protection that is fundamental to federated learning.

\subsection{Feature Replay for Mitigating Non-IID Distributions}

In the previous experiments, we demonstrated that the non-IID issue leads to significant performance degradation and slower convergence in FL. And this challenge has also been supported in \cite{zhao2018federated}, it demonstrates that sharing even a small portion of the dataset across clients can dramatically improve the global model performance by reducing gradient conflicts. However, while data sharing effectively mitigates non-IID effects, it comes at the cost of violating one of the fundamental principles of FL: privacy preservation. Directly exchanging raw images compromises data ownership and exposes sensitive information, making such approaches impractical for privacy-sensitive domains like agriculture.

In our framework, we take advantage of transfer learning to avoid these privacy risks. Specifically, we freeze the vision encoder of a pretrained CLIP model and only train a lightweight classifier over all clients. Since all clients share the same frozen encoder, the training input is not raw images but rather vision embeddings extracted from the CLIP encoder. This design provides a unique advantage: the encoder effectively acts as an encryption mechanism that transforms high-dimensional images into compact feature representations. Instead of considering direct image sharing, our method leverages these embeddings as the foundation for communication and collaboration among clients.

Importantly, these extracted features are non-reversible to raw images and thus do not compromise data privacy. In other words, while they are sufficient for classification tasks, they do not allow adversaries to recover the original images. This ensures that feature sharing aligns with the privacy requirements of FL, maintaining the integrity of sensitive agricultural data while still enabling cross-client collaboration.

To address the non-IID problem, our framework introduces a feature replay pool, formed by sharing a very small portion of extracted embeddings from all clients. During training, each client replays these shared embeddings alongside its private dataset, gaining exposure to representations of classes that may not exist locally. This mechanism helps align gradient updates toward the global optimum and reduces conflicts during aggregation. We will provide more detail about this technique in next section and the effectiveness of this approach will be evaluated in detail in the Results section. We will demonstrate that feature replay mitigates the negative impact of non-IID distributions while preserving the privacy guarantees of FL. The feature replay offering an efficient and secure solution for collaborative agricultural intelligence.

%% file: Sections/method.tex
\section{Methods}
\label{sec:methods}
In this section, we will provide details about the proposed FL framework. We will introduce the transfer learning classification model we select, and then introduce the workflow, at the end we will also discuss how our framework handle late-joining new client during training process.

\subsection{Transformer-based classification model with transfer learning}
\label{sec:classifier}

\begin{figure}[htp]
    \centering
    \includegraphics[width=1\linewidth]{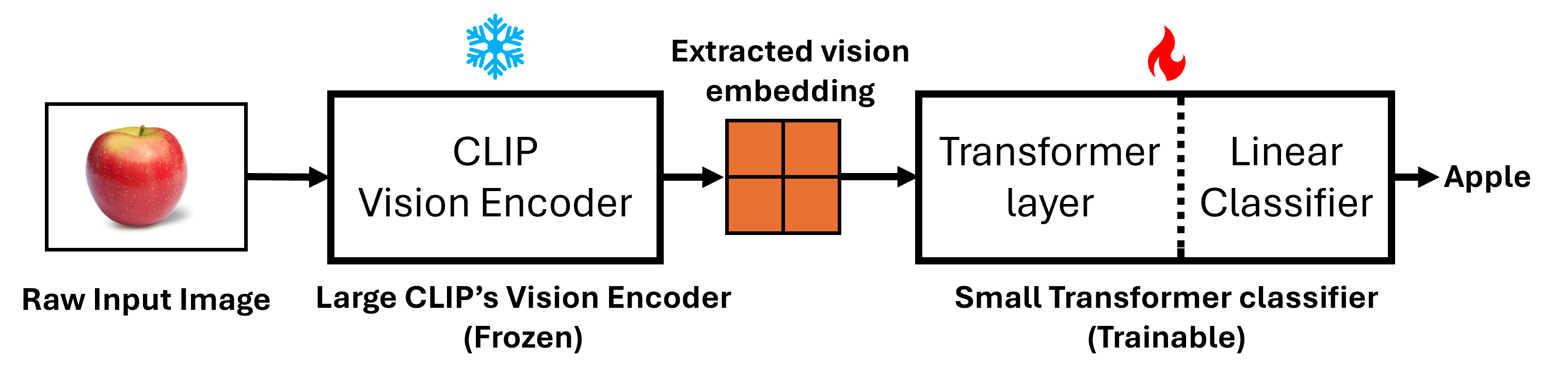}
    \caption{Structure of the proposed transformer-based classification model with transfer learning}
    \label{fig:classifier}
\end{figure}

To achieve efficient and high adaptability visual recognition under federated constraints, the classification model we designed adopts a transfer-learning strategy that leverages a frozen CLIP Vision Transformer (ViT-B/32) backbone as a universal feature extractor, and a lightweight Transformer-based classifier head for local adaptation. The structure of the proposed classification model with transfer learning is presented in Fig. \ref{fig:classifier}.  The CLIP ViT, pretrained on large-scale image–text pairs, provides robust and semantically aligned visual representations that generalize across diverse scenes. During the training, the framework avoids high communication overhead by freezing CLIP ViT parameters. This design enables each client to perform task-specific fine-tuning solely on a compact, trainable head, drastically reducing the number of synchronized parameters.

The Transformer classifier head is designed to refine the CLIP embeddings and capture lightweight contextual dependencies before classification. The lightweight classifier include two major components, a transformer encoder comprised by two transformer encoder layers used for feature embedding understanding and a linear layer used as classifier that produce classification output. The trainable parameters in this compact classifier constitute roughly 2\% of the total parameters in the entire model include the CLIP ViT. Which means, training such classifier in a FL can reduce 98\% of the communication overhead compared to training entire model from stratch. This modular design allows the model to adaptively transfer CLIP’s general vision knowledge to downstream tasks while maintain high accuracy and lower communication overhead.

During federated learning, only the parameters of the Transformer head are updated and transmitted between the server and clients, while the CLIP encoder remains fixed. This separation effectively implements federated transfer learning, where global semantic knowledge from CLIP is reused, and only lightweight model updates are collaboratively optimized. 

\begin{algorithm}[t]
\caption{FedReplay algorithm}
\label{alg:fedreplay}
\begin{algorithmic}[1]

\Require Number of communication rounds $R$, client number $N$, local epochs $E$, Global Embedding Exchange ratio $\lambda$, Global embedding replay pool $\mathcal{D}_{pub}$, client datasets $\{\mathcal{D}_i\}_{i=1}^N$,initialization of global transformer head $\theta^0$.

\vspace{0.5em}
\State \textbf{Server executes:}
\State Freeze CLIP encoder $f_{CLIP}$; initialize transformer head $f_\theta$ with parameters $\theta^0$.
\State Collect $\mathcal{D}_{pub}$ from all participant clients.
\State Pre-train $f_\theta$ on $\mathcal{D}_{pub}$ for $E_{warm}$ epochs as warm start.
\State Send $\mathcal{D}_{pub}$ to all client.
\For{$r \gets 1$ to $R$}
    \State Select a subset of clients $S^r \subseteq \{1, \ldots, N\}$.
    \For{$i \in S^r$}
        \State Receive locally updated parameters $\theta_i^{r}$ from client $i$ 
    \EndFor
    \State Aggregate the model parameters:
    \[
        \theta^{r+1} \gets \frac{1}{|S^r|} \sum_{i \in S^r} \theta_i^{r+1}
    \]
    \State Send current global model $\theta^{r+1}$ to all clients in $S^r$.

\EndFor

\vspace{0.1em}
\noindent\hrulefill
\vspace{0.1em}

\State \textbf{Client $i$ executes:}

\State Receive global embedding replay pool $\mathcal{D}_{pub}$ from server.
\State Receive model parameters $\theta^r$ from server.
\State Build replay buffer $\mathcal{R}_i$ by sampling balanced features from $\mathcal{D}_{pub}$ for missing classes.
\For{$e \gets 1$ to $E$}
    \State Sample a original data batch $x$ from $\{\mathcal{D}_i\}$
    \State Extract frozen CLIP features $z = f_{CLIP}(x)$.

    \State Compute local classification loss $\mathcal{L}_{local}$.

    \State Sample replay embedding batch $y$ from $\mathcal{R}_i$.
    \State Compute replay loss $\mathcal{L}_{replay}$.
    \State Combine losses:
    \[
        \mathcal{L} = (1 - \lambda)\mathcal{L}_{local} + \lambda \mathcal{L}_{replay}
    \]
    \State Update parameters from $\theta_i^{r+1} \gets \text{AdamW}(\nabla_\theta \mathcal{L})$.

\EndFor
\State Send updated $\theta_i^{r+1}$ to server if been selected.
\end{algorithmic}
\end{algorithm}
\vspace{-0.5em}

\subsection{Overall workflow of proposed FL framework}

\begin{figure*}[htp]
    \centering
    \includegraphics[width=1\linewidth]{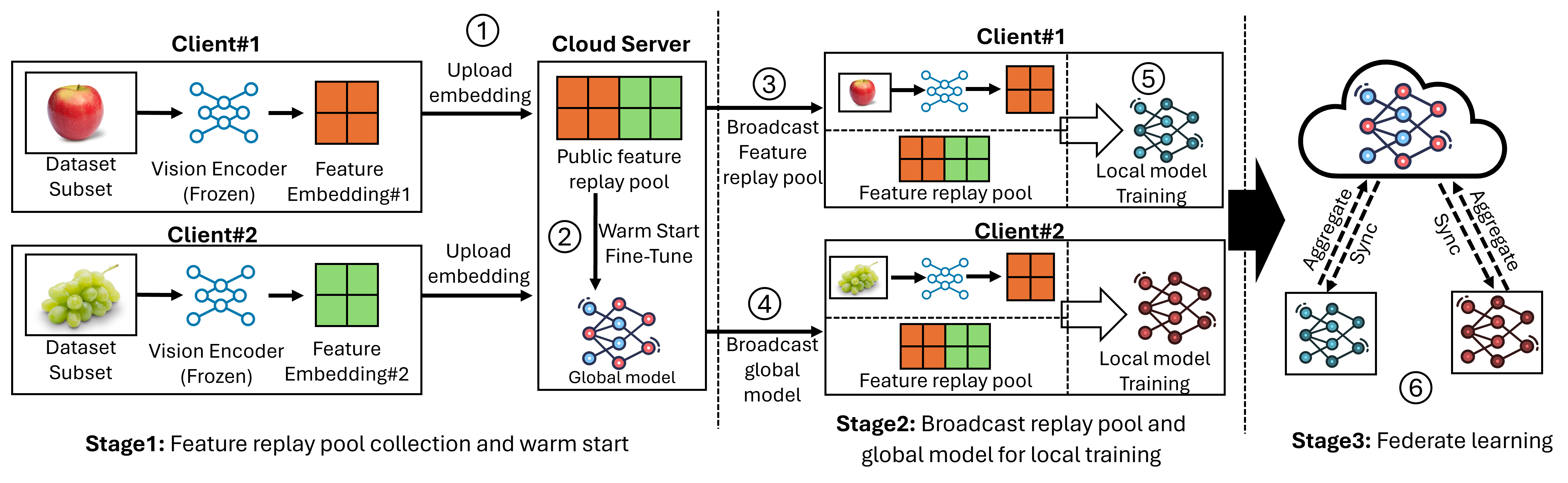}
    \caption{Workflow of training process in the proposed FL framework}
    \label{fig:overall_workflow}
\end{figure*}

The transfer-learning-based Transformer classifier implemented in our FL framework significantly reduces communication overhead during the FL communication rounds, thereby mitigating one of the critical challenges faced by conventional FL approaches, as discussed previously. However, FL is still is affected by the non-IID data issue, where the aggregated gradients from heterogeneous clients may conflict with each other, leading to a degradation in the global model’s accuracy, as demonstrated in Section.~ \ref{sec:non-iid}. In this section, we introduce the detailed workflow of the proposed FL framework and demonstrate how it addresses the non-IID problem to enhance overall performance. The workflow of the proposed FedReplay framework is illustrated in Fig.~ \ref{fig:overall_workflow}. The Algorithm. \ref{alg:fedreplay} demonstrate the detail process of the entire federated learning communication.

\subsubsection{Feature replay pool collection}
In our classification model demonstrate in Fig. \ref{fig:classifier}, the client first extract vision feature embedding from private input images, and then use a transformer classifier to perform classification on the extracted embedding. The frozen CLIP ViT is identical across all client, which can be considered as a data encryption encoder. The extracted feature embeddings are non-reversible to raw images and thus do not compromise data privacy.

Based on this feature, we propose feature replay strategy to address non-IID issue. Feature replay allow data share across clients while maintain data privacy. As demonstrate in Fig. \ref{fig:overall_workflow}, all client will sample a small subset from its private dataset and extracted feature embedding from it. Then all client upload the feature embedding subset to the server. The server will combine the feature embedding received from all client as a public feature replay pool which include extracted feature embedding from all class across clients. This feature replay pool can provide representations of classes that may not exist for a client, and helps align gradient updates toward the global optimum and reduce the conflicts caused by non-IID issue.

\subsubsection{Warm start fine-tune on server}

Before initiating distributed federated optimization, the server performs a warm-start fine-tuning of the global model using the shared feature replay pool. This replay pool contains a small, balanced subset feature embeddings representing global class distribution. Fine-tuning the initialized global Transformer classifier on this replay pool allows the model to acquire a balanced decision boundary and a globally aligned embedding space prior to client-specific training. This process effectively avoid the common cold start problem common in federated learning to help metigate non-IID issue, where randomly initialized classifiers struggle to converge under non-IID data distributions. 

By pre-adapting the model on globally representative features, the warm start reduces early-round instability, accelerates convergence, and provides a fairer initialization for all participating clients. Moreover, this initialization enables the subsequent local models to focus on learning domain-specific variations rather than re-establishing global class separability, leading to faster and more stable collaborative optimization. After warm start, the server will broadcast the feature replay pool and fine-tuned global model weight to all client and start the feaderated learning process.

\subsubsection{Local training with feature replay}

After obtain the global model weight and feature replay pool, the local client will initialize the local model by global model weight and start local training process. As shown in Algorithm. \ref{alg:fedreplay}, the local client training include two learning process: 1) conventional training over private local dataset, 2) feature replay training over feature replay pool. 

The client will first sample a batch of data $(data_L,label_L)$ from its private dataset and extract feature embedding batch $(embedding_L,label_L)$, then feed it to the local classifier model to get final result. At the end the client will compute local classification loss $\mathcal{L}_{local}$ using equation \ref{eq:local_loss}:
\begin{equation}
\mathcal{L}_{\text{$local$}} = \mathrm{CE}\big(f_{\theta}(embedding_L),\, label_L\big)
\label{eq:local_loss}
\end{equation}

The $\mathrm{CE}$ is cross entropy function and $f_{\theta}$ is the transformer classification model. The $\mathcal{L}_{local}$ represent the update gradient that model purely learned from local dataset that doesn't contain class sample from other clients. Conventional FL rely on this loss to update model so that lead to conflict gradient and degrade performance cause by non-IID issue.

Therefore, we conduct feature replay learning process alongside with the normal local learning process to help align the gradient update toward global optimal direction. After local learning, each client will also sample a batch of embedding $(embedding_R,label_R)$ from feature replay pool. The sampled data is already feature embedding so can be directly feed into transformer classifer model to get result and calculate a replay loss $\mathcal{L}_{replay}$:
\begin{equation}
\mathcal{L}_{\text{$replay$}} = \mathrm{CE}\big(f_{\theta}(embedding_R),\, label_R\big)
\label{eq:replay_loss}
\end{equation}

After that the cient will combine two loss via equation \ref{eq:total_loss} to get final losses and use it for back propagation update of the compact transformer classifier.

\begin{equation}
\mathcal{L} = (1 - \lambda)\,\mathcal{L}_{\text{local}} 
              + \lambda\,\mathcal{L}_{\text{replay}},
\label{eq:total_loss}
\end{equation}

The $\lambda$ is used to control replay ratio that balance the knowledge learned from  imbalance private dataset and balanced feature replay pool. This mixture is essential because it aligns class distributions without overwriting the client’s private data distribution.

\subsubsection{Federated learning aggregation}

The global model is collaboratively training over clients through federated learning as illustrate in Fig. \ref{fig:overall_workflow}. In each communication round, a subset of clients will be randomly selected, denoted by $S^r$, to participant model aggregation. These clients will send their local model weight to the server. The server will aggregate all model weight and sync back the aggregated model weight back those selected clients $S^r$. This process will repeat several rounds untill the global model converged. The full process of the federated learning is described in Algorithm. \ref{alg:fedreplay}.

\subsection{Late-joining client integration}

In practical FL deployment, participating clients rarely remain static throughout the training lifecycle, it must operate over dynamic and heterogeneous client networks, where new farms, sensors, or data collection devices are continuously added as the system expands. The agricultural environments evolve over time, new crop species, pest types, or growth conditions emerge, and different institutions or regions may join the federation after the initial training phase. Consequently, the ability to integrate late-joining clients becomes indispensable for building sustainable and scalable federated agricultural intelligence systems. 

However, current existing FL framework are typically designed under the assumption of a fixed client set and static label space. When a new farm client contributes previously unseen crop varieties or disease categories, the global model often fails to incorporate this knowledge without full retraining. This lead to two critical issues: (1) \textbf{Catastrophic forgetting and representation misalignment:} where Incorporating new data distributions or novel classes causes degradation of performance on previously learned tasks. (2) \textbf{Communication inefficiency:} since retraining or re-synchronizing large-scale models such as vision transformers across all clients is prohibitively costly under limited rural or edge-network bandwidth.

\begin{figure*}[htp]
    \centering
    \includegraphics[width=1\linewidth]{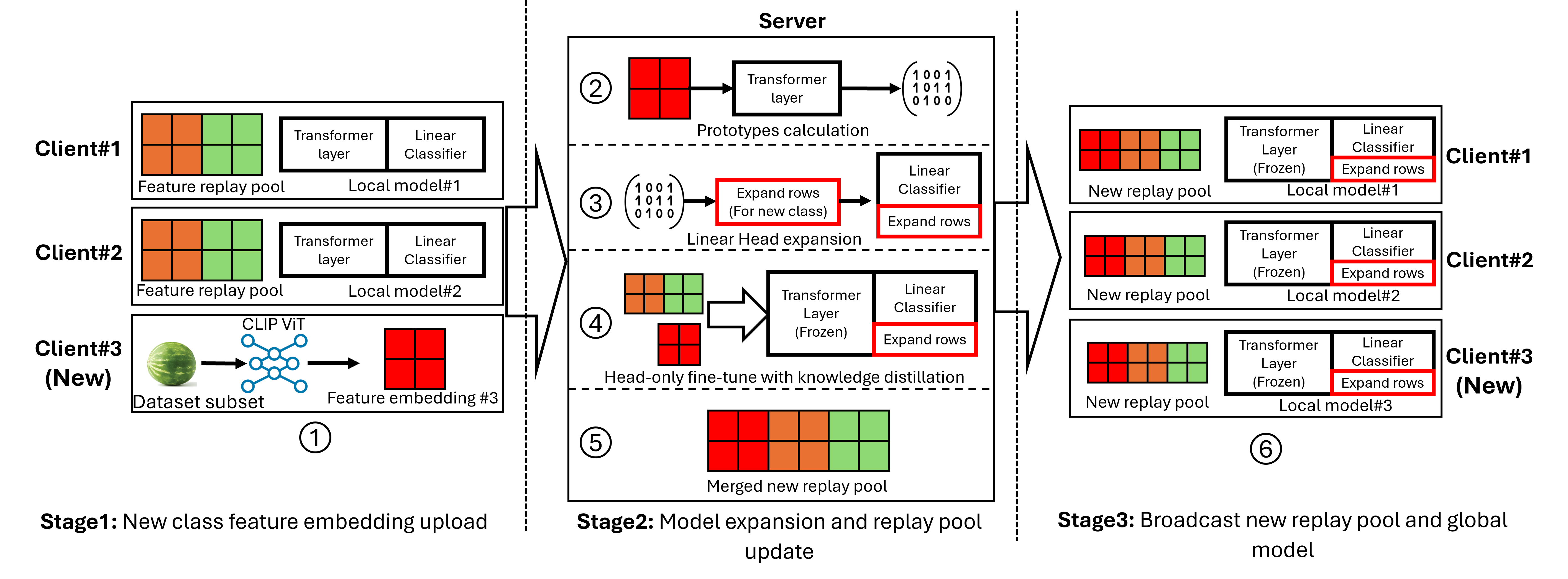}
    \caption{Workflow of the late-joining client integration process in the proposed FL framework}
    \label{fig:workflow_newclient}
\end{figure*}

In our proposed FL framework, we take this issue into consideration. In Fig. \ref{fig:workflow_newclient}, we demonstrate the mechanism we implemented to integrate late-joining new client. For illustration purpose, we assume there are already two exist clients (Client 1 and 2) in the system and the training process is finished. Laterly new client (Client 3) that include new unseen class want to join the FL system.

\subsubsection{Model expansion and replay pool update}

To accept the new client, current FL system need to expands the classifier model and update the replay pool across client to include new class. The new client need to sample a subset from its local private dataset and extract feature embedding from it. Then the new client will upload the extracted feature embedding pool to the server.

When server received embedding pool from new client, it first computes a set of class prototypes to characterize these new categories in the shared feature space. The prototypes is the average output of transformer encoder before linear classifier. Formally, for new class $c$, it's prototype is computed as:

\begin{equation}
\mathbf{p}_c = \frac{1}{N_c} 
\sum_{i=1}^{N_c} 
\frac{Trans(feature)}{\|Trans(feature)\|},
\label{eq:prototype}
\end{equation}

Where $Trans$ denote the transformer encoder in classifier model and $N_c$ is the number of sample. The $feature$ is the CLIP extracted feature from sample subset. The resulting feature vectors are normalized and averaged within each class to form class-level prototypes that represent the semantic center of the new category in the latent space. These prototypes effectively capture the average visual representation of each new class and provide an informed initialization for model expansion to avoid randomly initialization.

Once the prototypes are obtained, the server expands the global linear classification layer to accommodate the new classes. The existing classifier contains $C_{old}$ output neurons corresponding to the previously learned categories. For each new class, a new row is appended to the classifier’s weight matrix, and its parameters are initialized using the corresponding prototype vector:

\begin{equation}
\mathbf{W}_{C_{\text{old}}+c} = \mathbf{p}_c, 
\quad 
\mathbf{b}_{C_{\text{old}}+c} = 0,
\label{eq:expansion}
\end{equation}

Because both the prototypes and classifier weights exist in the same normalized latent space, this initialization directly aligns the new decision vectors with the actual feature distributions of the newly introduced classes. To be noticed, the Transformer encoder does not require retraining or structural modification during this process, only the classifier head must be extended to map expanded new label set. This selective expansion allows the model to learn new knowledge efficiently while retaining the pretrained visual reasoning capabilities of the frozen encoder.

Once the classifier head is is expanded to include new classes, naively continuing training induces a stability–plasticity conflict: learning the new classes (plasticity) can overwrite decision boundaries for existing classes (stability), causing catastrophic forgetting and breaking backward compatibility for clients already deployed. But full end-to-end retraining would be communication-heavy and would shift the shared embedding space, forcing every client to re-adapt from scratch. To resolve this issue, we implement a head-only fine-tuning procedure with knowledge distillation (KD) \cite{li2024continual, kang2022class}. A frozen, pre-expansion global model serves as the teacher, supplying soft targets that encode inter-class structure and calibration on the original label set.  While the student, the expanded model with the same frozen encoder, learns the new classes via supervised loss and constrains its responses on old classes to match the teacher. 

During tuning, the student learns to predict new classes using cross-entropy loss on the new feature replay pool, while simultaneously align its response matrix on the old classes by imitate the teacher’s softened logits. This procedure adjusts only the linear classifier parameters, keeping the Transformer encoder frozen to maintain a consistent feature representation across clients. This design preserves prior knowledge, prevents catastrophic forgetting, and limits communication overhead by updating only the linear classifier while keeping the Transformer encoder fixed. Formally, the total loss for head-only fine-tune with KD is formulated as:

\begin{equation}
\mathcal{L} 
= 
\mathcal{L}_{\text{CE}}^{\text{new}} 
+ 
\lambda_{\text{KD}}
\mathrm{KL}\!\left(
p_{teacher}^{(\text{old})}
\,\|
\,p_{student}^{(\text{old})}
\right),
\label{eq:kd_loss}
\end{equation}

In which $\mathcal{L}_{\text{CE}}^{\text{new}}$ is the cross entropy loss computed on new feature replay pool, $\mathrm{KL}(\|)$ is used to calculate the Kullback–Leibler divergence of  probability distributions over the old classes between teacher model $p_{teacher}^{(\text{old})}$ and student model $p_{student}^{(\text{old})}$. $\lambda_{\text{KD}}$ is the temperature parameter used to balance the loss from new class and knowledge aligned with teacher.

The total loss $\mathcal{L}$ is then used to update the expanded model. As a result, the expanded model inherits the knowledge of old classes from the teacher while adapting to new categories introduced by the late-joining client, achieving stable and backward-compatible integration without retraining the entire network.

\subsubsection{New client transition and resume FL learning}

After model expansion and head-only fine-tune with KD, the server will update the public feature replay pool to include the new class feature embedding. Then, the server will broadcast the new feature replay pool and expanded global model back to all client. The new client now is activated and become part of the system member.

However, if we resume the normal FL process immediately, all client, including those without data for the new classes, would contribute updates to the entire classifier matrix. The newly added weight rows will fluctuate unpredictably and potentially destroy the semantic consistency established during prototype initialization and knowledge distillation. Because the model weight is already align with the old class, and new client who responsible for new unseen class will not have sufficient time to contribute optimal weight update for expanded rows since it will be submerged by all other client. This uncontrolled parameter mixing often leads to gradient noise, slower convergence, and severe degradation in recognition accuracy for both old and new categories.

To address this problem, the proposed framework introduces Row-Gated Federated Averaging (Row-Gated FedAvg), a selective aggregation strategy that isolates parameter updates based on class ownership. During the transition period following model expansion, the server aggregates expanded row of the linear classifier only from those clients that possess local data for the new class. The Row-Gated FedAvg will only be applied on linear classifier while transformer encoder parameters are aggregated via standard FedAvg.

This row-wise gating mechanism ensures that newly introduced classes evolve only under supervision from clients that truly observe them, preserving the stability of previously learned representations. By preventing meaningless gradient interference from irrelevant clients, Row-Gated FedAvg provides a controlled adaptation stage where the new class weights can mature before full federation resumes. Once the new weights stabilize, gating is lifted and normal FedAvg aggregation proceeds across all parameters.

%% file: Sections/result.tex
\section{Experiment}
\label{sec:result}
We conduct a comprehensive evaluation to evaluate the efficiency of the proposed FL framework. In this section, we demonstrate the comparison result under non-IID data distribution and evaluate the impact of proposed techniques implemented.

\subsection{Experiment Setup}
\subsubsection{Dataset}
The dataset we used to conduct evaluation test is CWD30 \cite{ilyas2025cwd30}. The CWD30 dataset is a large-scale and holistic benchmark designed for crop–weed recognition in precision agriculture. It contains more than 219,000 high-resolution images covering 10 crop species and 20 weed species, captured across multiple growth stages, viewing angles, and environmental conditions. Unlike earlier agricultural datasets that are often limited in scope, CWD30 offers a hierarchical taxonomy and diverse real-field scenarios, enabling fine-grained classification and robust model training. Its comprehensive coverage of intra-class variability and inter-class similarity makes it particularly challenging and realistic, providing a valuable resource for developing and evaluating deep learning models in crop–weed detection and agricultural intelligence.

\subsubsection{Experiment Setting}

We conduct an comprehensive evlaution on a workstation with a AMD Ryzen Threadripper PRO 3955WX 16-Cores CPU and a NVIDIA RTX A6000 GPU. The key parameter we used in proposed FL framework are presented in Table \ref{tab:fl_parameters}. 

\begin{table}[ht]
\centering
\caption{Key parameters used in the proposed FL framework.}
\label{tab:fl_parameters}
\renewcommand{\arraystretch}{1.2}
\begin{tabular}{lcl}
\toprule
\textbf{Parameter} & \textbf{Symbol / Value} \\
\midrule
Number of clients & $5$   \\
Total number of classes & $30$  \\
Communication rounds & $500$  \\
Local epochs per round & $1$  \\
Batch size & $256$  \\
Learning rate & $1\times10^{-3}$  \\
Weight decay & $1\times10^{-4}$  \\
Replay ratio & $\lambda = 0.5$  \\
Optimizer & AdamW \\
Loss & Cross-Entropy \\
Public feature ratio & $1\%$  \\
Transformer encoder depth & $L = 2$  \\
Frozen CLIP encoder & ViT-B/32  \\
Trainable parameters & $1.8$M ($\sim$2\% of CLIP) \\
Total parameters & $88.8$M & \\
KD loss temperature & $\lambda_{\text{KD}} = 0.5$  \\
Row-gated rounds & $R_{\text{gate}} = 10$  \\
Warm-start epochs & $E_{\text{warm}} = 5$  \\
\bottomrule
\end{tabular}
\end{table}

We assume a FL system that include 5 clients, the number of unique class assigned to each client is 6. The FL communication will last 500 rounds, and each client only contribute 1\% of its own feature embedding to public feature replay pool. Before training, the server will perform 5 epochs warm start fine-tune on public feature replay pool.

\subsubsection{Baseline Methods}
In our evaluation, we compare our proposed method with some baseline method in below to verify the efficiency and effectiveness.
\begin{itemize}
    \item Raw-CLIP: Directly applies the pretrained CLIP model without fine-tuning. This setting reflects the out-of-the-box capability of CLIP on agricultural classification tasks and serves as a lower-bound reference.
    \item FedTPG: The method proposed in \cite{qiu2024federated}, which freezes the CLIP backbone and trains a text prompt generator collaboratively in FL to adapt CLIP for target domains.
    \item StdFed: A baseline variant of our framework that trains a lightweight transformer classifier on CLIP-extracted embeddings in a standard FL setting. Unlike our proposed method, it does not include feature replay and therefore suffers from non-IID performance degradation.
    \item Centralized Fine-Tuning: A centralized training setting where the classifier is optimized on the full dataset with access to all samples. This serves as an upper bound for evaluating the effectiveness of federated methods.
\end{itemize}
\subsection{Evaluation Result}

\begin{table*}
\centering
\caption{Aggregated model per-class accuracy comparison across different methods (\%).}
\label{tab:accuracy}
\input{Comparison_Table}
\end{table*}

\subsubsection{Baseline Comparison}

\begin{figure}[htp]
    \centering
    \includegraphics[width=0.85\linewidth]{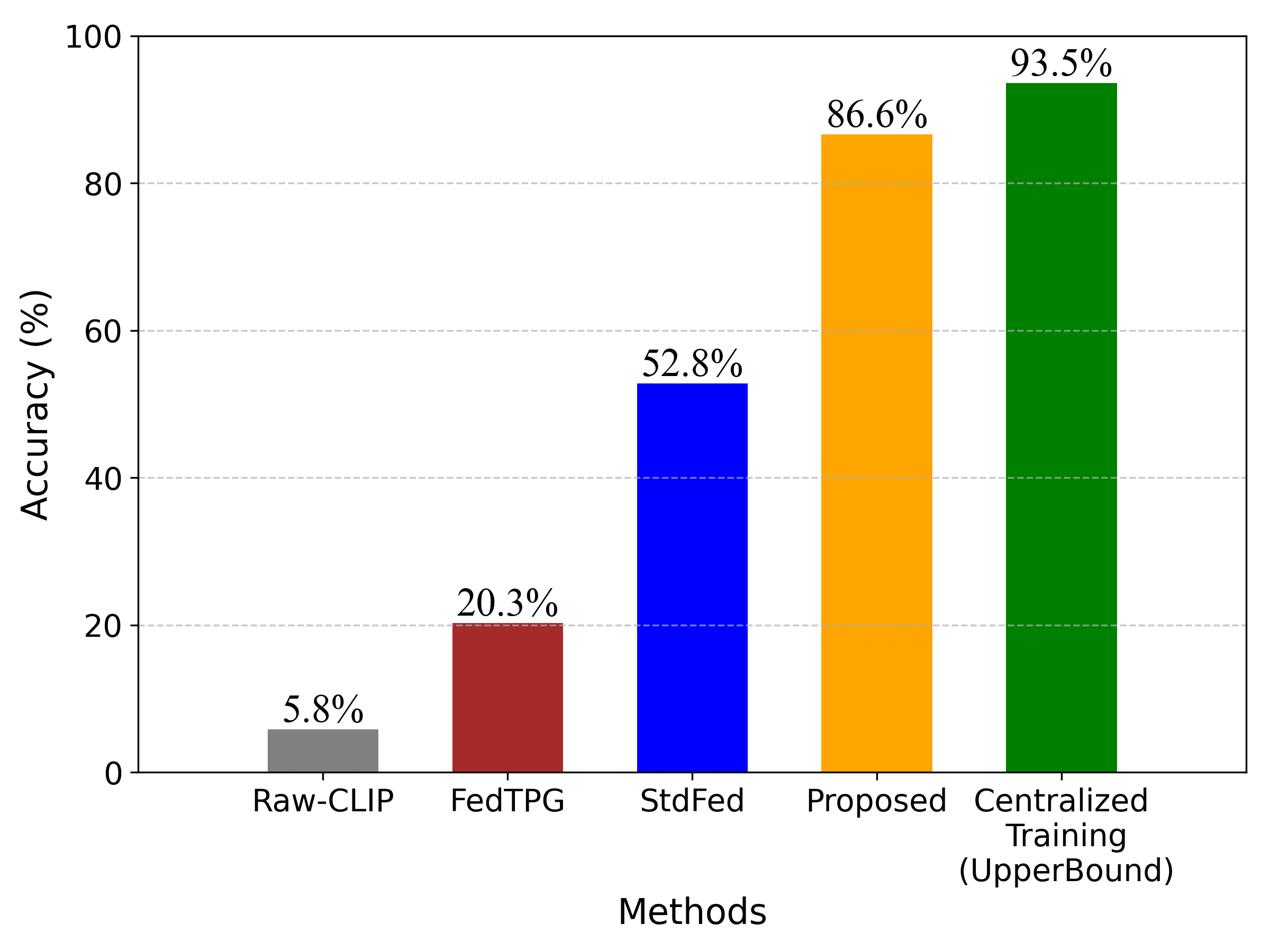}
    \caption{Performance comparison}
    \label{fig:comparison}
\end{figure}

Fig. \ref{fig:comparison} presents the performance comparison between our proposed method and several baseline approaches. The centralized fine-tuning result serves as the upper bound, achieving an accuracy of 93.5\%, which demonstrates the high potential of transfer learning with a CLIP-based vision encoder when trained on the full dataset. This establishes the theoretical ceiling of our framework under ideal centralized conditions.

The baseline Raw-CLIP model, which directly applies the pretrained CLIP without fine-tuning, performs poorly with only 5.8\% accuracy. This result highlights that while CLIP encoders are powerful on general-purpose datasets, they lack the specialized agricultural knowledge required for crop and weed recognition tasks. FedTPG, which incorporates text-driven prompt generation to adapt CLIP in a federated setting, improves accuracy to 20.3\%. Although prompt learning provides additional textual cues, the method is still heavily constrained by the non-IID distribution of data across clients, and the benefit from prompt information alone is limited.

StdFed, which trains a lightweight transformer classifier in traditional FL setting achieves 52.8\% accuracy, outperforming FedTPG. This improvement comes from directly learning task-specific features rather than relying solely on textual prompts. However, as discussed in the previous section, tradition FL suffers from the non-IID issue, This result confirms that FL on agricultural datasets faces significant convergence and performance barriers.

Our proposed method, which introduces feature replay into the StdFed framework, achieves a substantial performance boost, reaching 86.6\% accuracy. By sharing a small subset of non-reversible embeddings across clients, the replay pool provides balanced exposure to diverse class information, effectively mitigating the negative effects of non-IID distributions. Compared with FedTPG, our method delivers more than a 4 times improvement and relative to StdFed, it demonstrates that feature replay is crucial for aligning local updates with the global optimum. While a 7\% gap remains between the proposed method and the centralized upper bound, this can be attributed to the inherent knowledge loss in federated learning, which is the trade-off for ensuring data privacy. In addition, our framework freezes the CLIP vision encoder and only trains the lightweight transformer classifier, the trainable component accounts for just 2\% of the total parameters, reducing communication overhead by approximately 98\% compared to training the entire framework from scratch. This demonstrates that our design not only achieves high accuracy but also significantly reduce the communication overhead in FL.

To provide further insight,  Table \ref{tab:accuracy} presents the per-class accuracy of each method across the 30 crop weed categories in the CWD30 dataset. The results reveal consistent improvements of the proposed framework over both FedTPG and StdFed across nearly all classes. Notably, in difficult cases such as asiatic-dayflower, livid-pigweed, and korean-dock, our approach demonstrates accuracy gains exceeding 30\% compared to StdFed. These findings confirm that the feature replay mechanism effectively reduces the bias introduced by non-IID local datasets, leading to a more balanced model capable of generalizing across diverse crop and weed species.

\subsubsection{Impact of Participant Rate on Federated Learning}

\begin{figure}[htp]
    \centering
    \includegraphics[width=0.75\linewidth]{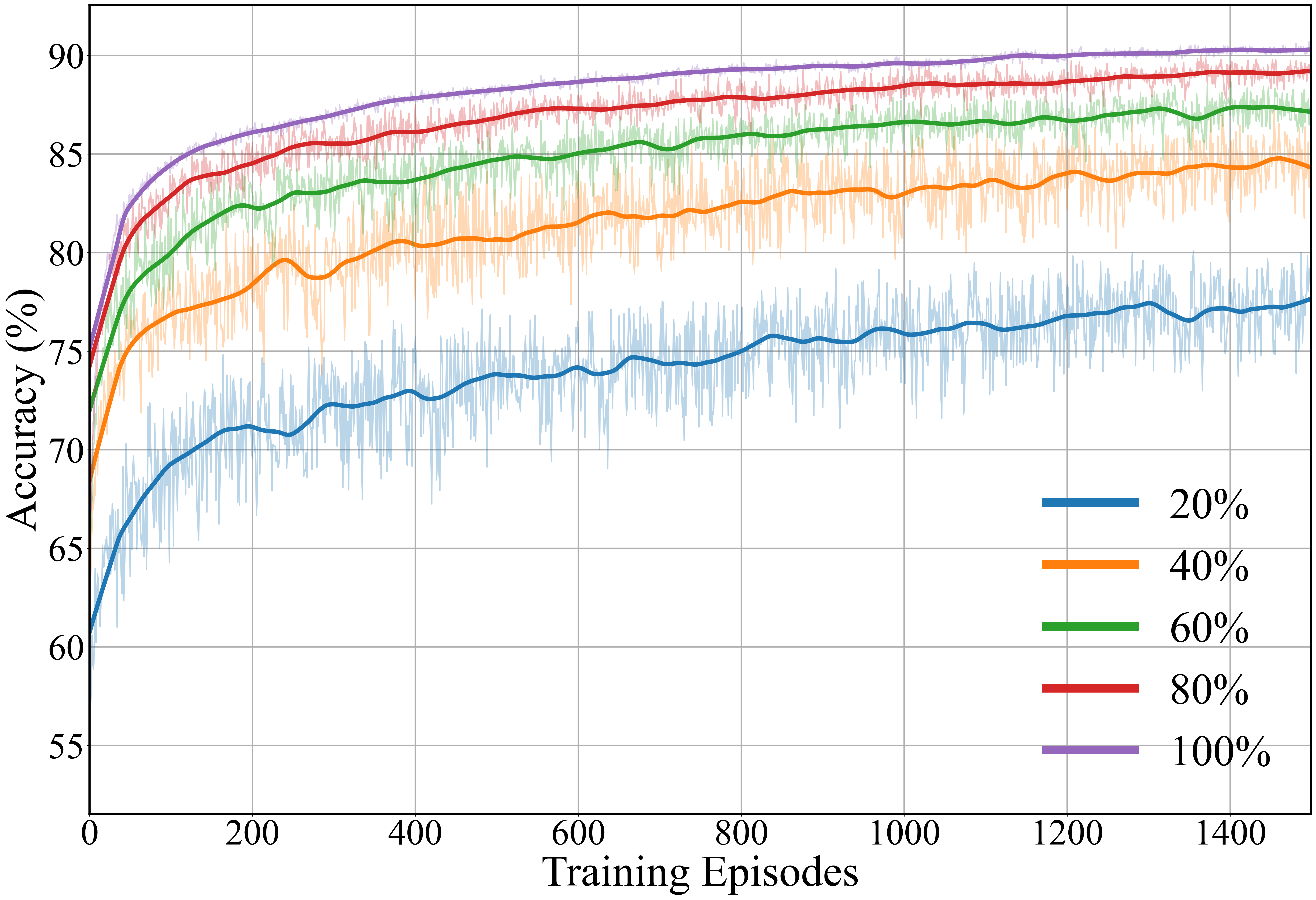}
    \caption{Validation accuracy over different Participant Rate}
    \label{fig:participant_rate}
\end{figure}

In this experiment, we examine how the participant rate in federated learning influences both the training dynamics and the final model performance. The participant rate determines the fraction of clients that are randomly selected to participate in each communication round, thereby governing the degree of update diversity and representativeness in the global aggregation process. A higher participation rate typically enhances the stability of convergence by incorporating more heterogeneous local gradients, whereas a lower rate may accelerate training per round but risks introducing bias and slower global convergence due to limited client coverage. To systematically evaluate these trade-offs, we conduct a controlled experiments under different participation rates and monitor their effects on accuracy evolution across rounds. The results are illustrated in Fig. \ref{fig:participant_rate}, where both the raw accuracy trajectories and their LOWESS-smoothed curves are plotted to emphasize the overall convergence patterns and performance trends.

The results show that higher participant rates consistently lead to faster, more stable, and more accurate convergence. With only 20\% participation, the model not only converges to a lower final accuracy (below 75\%) but also requires many more episodes to approach this performance. This is because each round aggregates knowledge from only a small subset of clients, providing limited and biased updates. The learning curve also exhibits large fluctuations, reflecting unstable optimization. Increasing the participant rate to 40\% improves both accuracy and convergence speed, though instability remains visible. At 60\% and 80\% participation, the model benefits from richer updates, achieving faster convergence, higher final accuracy, and reduced fluctuation. Finally, training with 100\% participation achieves the best performance ( near 90\% validation accuracy), converges the fastest, and exhibits the smoothest trajectory with minimal oscillation.

This experiment reveals the trade-off between communication cost and performance in federated learning. Higher participation rates reduce stochasticity,accelerates convergence and improves stability at the expense of higher overhead. Lower participation reduces communication cost per round but slows convergence and amplifies instability. In our evaluation, we set the default participant rate to 60\%, which provides a balanced compromise between maintaining strong accuracy and controlling communication overhead, making it suitable for scalable agricultural FL deployment.

\subsubsection{Different Client Number}
\label{subsubsection:client_num}
\begin{figure}[htp]
    \centering
    \includegraphics[width=0.75\linewidth]{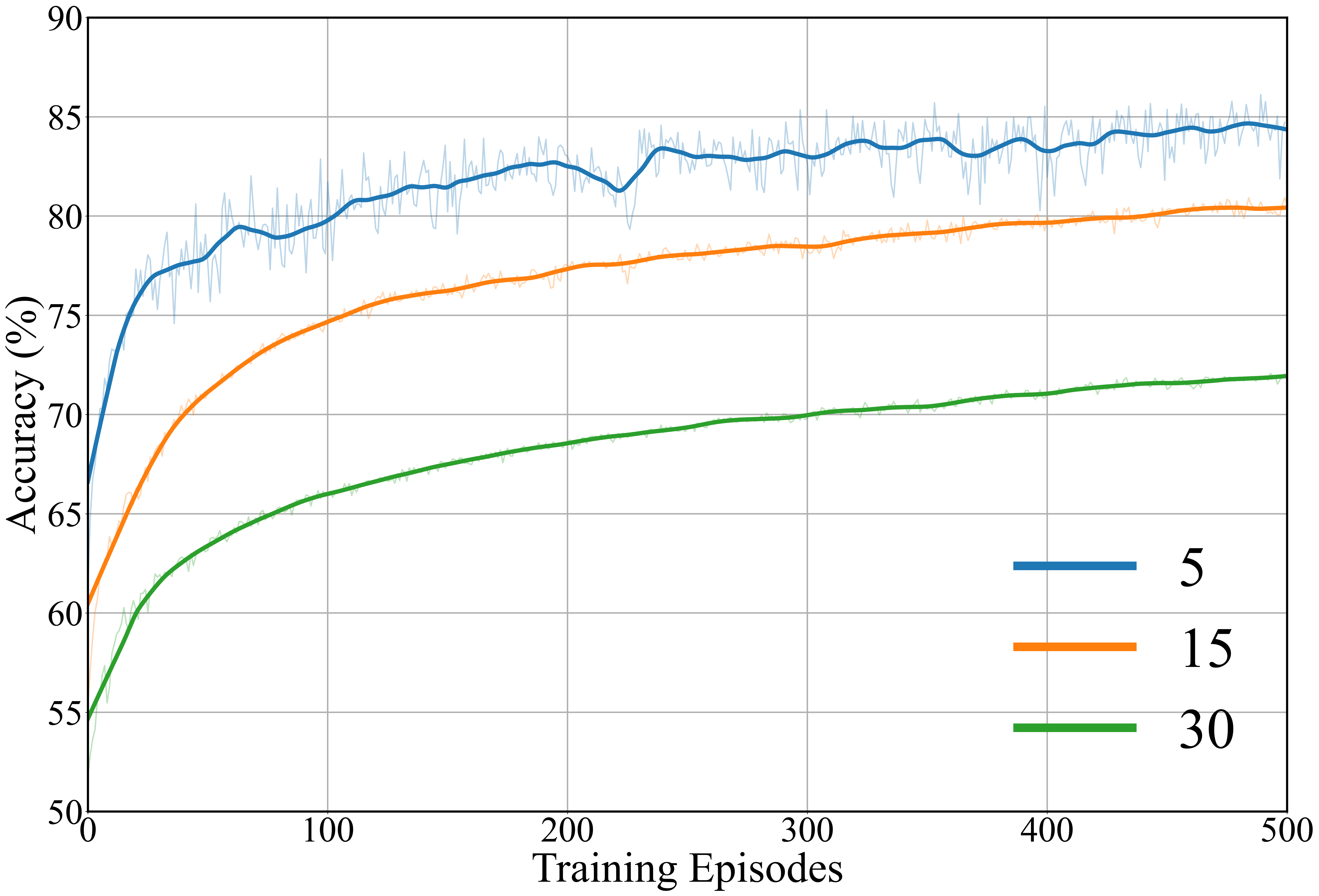}
    \caption{Validation accuracy over different number of client}
    \label{fig:client_num}
\end{figure}

In this experiment, we investigate how the total number of clients in the federated learning system affects both the training dynamics and the final model performance. We evaluate three configurations involving 5, 15, and 30 clients, respectively, while keeping the participant rate constant to isolate the effect of client population size. Increasing the number of clients introduces greater data diversity and communication sparsity, which may alter convergence behavior and aggregation stability. The 30 classes in the dataset are partitioned and assigned across all clients, with each client owning a non-overlapping subset containing only its designated classes. The results are presented in Fig. \ref{fig:client_num}, where both the raw accuracy trajectories and their LOWESS-smoothed curves are plotted.

From the result, larger client numbers negatively affect both convergence speed and final accuracy. With 5 clients, the model converges rapidly and achieves the highest accuracy of around 85\%, as each client retains relatively diverse data that better approximates the global distribution. When expanded to 15 clients, performance drops to about 80\%, and convergence becomes slower. In this setting, per-client datasets shrink and the between-client label disparity intensifies, producing noisier and more biased updates. If increase the client number to 30, the model convergence is the slowest and the final accuracy comes to 72\%. The federation now operates in a highly fragmented regime, each client only responsible for only one class. the partial participation in each round limits the effective label coverage seen by the server, and the replay pool’s class representation becomes thinner per class, all of which elevate aggregation variance and depress the convergence speed and final performance.

The result of this test confirm that excessive  fragmentation of dataset exacerbates gradient conflicts during aggregation and hinders global optimization. Increasing the client number under non-IID data distribution degrades both convergence speed and final performance. In this test, we select 5 client as default setting.

\subsubsection{Different feature replay share rate}

\begin{figure}[htp]
    \centering
    \includegraphics[width=0.75\linewidth]{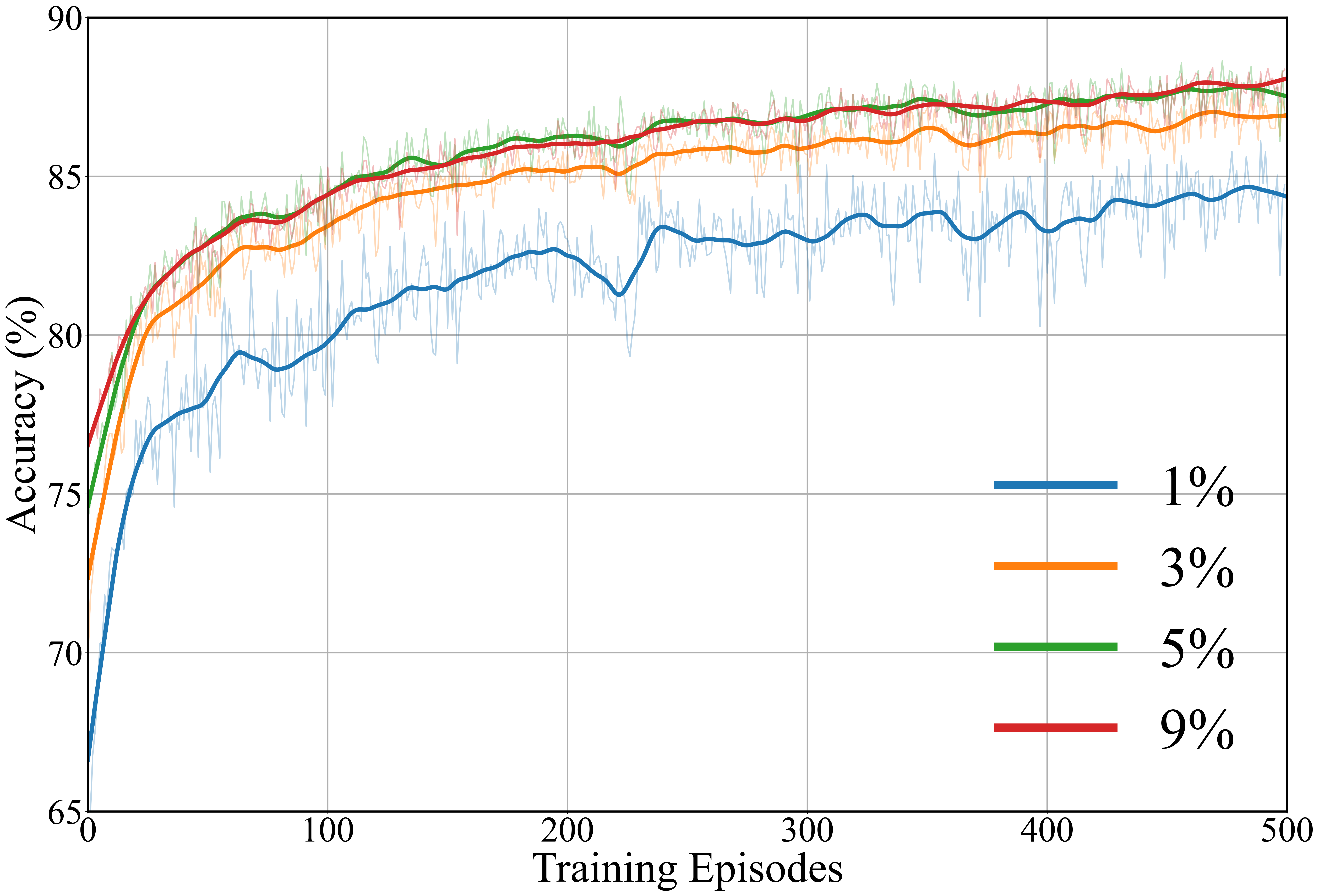}
    \caption{Validation accuracy over different feature shared rate}
    \label{fig:share_rate}
\end{figure}

In our proposed method, we implement feature replay to metigate non-IID issue in FL. We demonstrate in previous test, the performance will increased significantly by sharing only 1\% of the extracted feature among all clients. In this test, We investigate how the proportion of shared feature in the replay pool influences learning under non-IID data. All settings are fixed (frozen CLIP encoder, training lightweight transformer head, 60\% participation), only the share rate varies across 1\%, 3\%, 5\%, and 9\%. The raw accuracy trajectories and their LOWESS-smoothed curves under different feature share rate are presented in Fig. \ref{fig:share_rate}.

When share only 1\% extracted feature, accuracy rises markedly relative to no replay but the trajectory remains visibly noisy and the curve bends upward more slowly. With lower share rate, limited cross-client exposure reduces gradient conflict but does not fully anchor client updates to a common class prior, so aggregation retains higher variance. When raising the share rate to 3\% yields faster early gains and a smoother curve. The replay pool now covers more inter-class variation, and local models receive stronger signals about classes absent from their private data. Moving to 5\% brings the curve close to saturation, convergence is stable and the final accuracy approaches the best observed in this ablation test, indicating that the pool is sufficiently diverse to counter most non-IID drift. Increasing further to 9\% offers only marginal improvement; once the replay features span the essential class structure, additional samples add redundancy rather than new information while incurring extra bandwidth and storage.

In this test, we demonstrate only 1\% of shared feature already delivers a substantial boost.  3–5\% provides a strong accuracy–stability sweet spot with minimal overhead.The diminishing returns beyond 5\% suggest that our privacy-preserving design achieves most of its benefit without large exchanges, reinforcing the practicality of replay as a communication-efficient remedy for non-IID federated learning.

\subsubsection{Continual integration of new clients}

\begin{figure}[htp]
    \centering
    \includegraphics[width=0.75\linewidth]{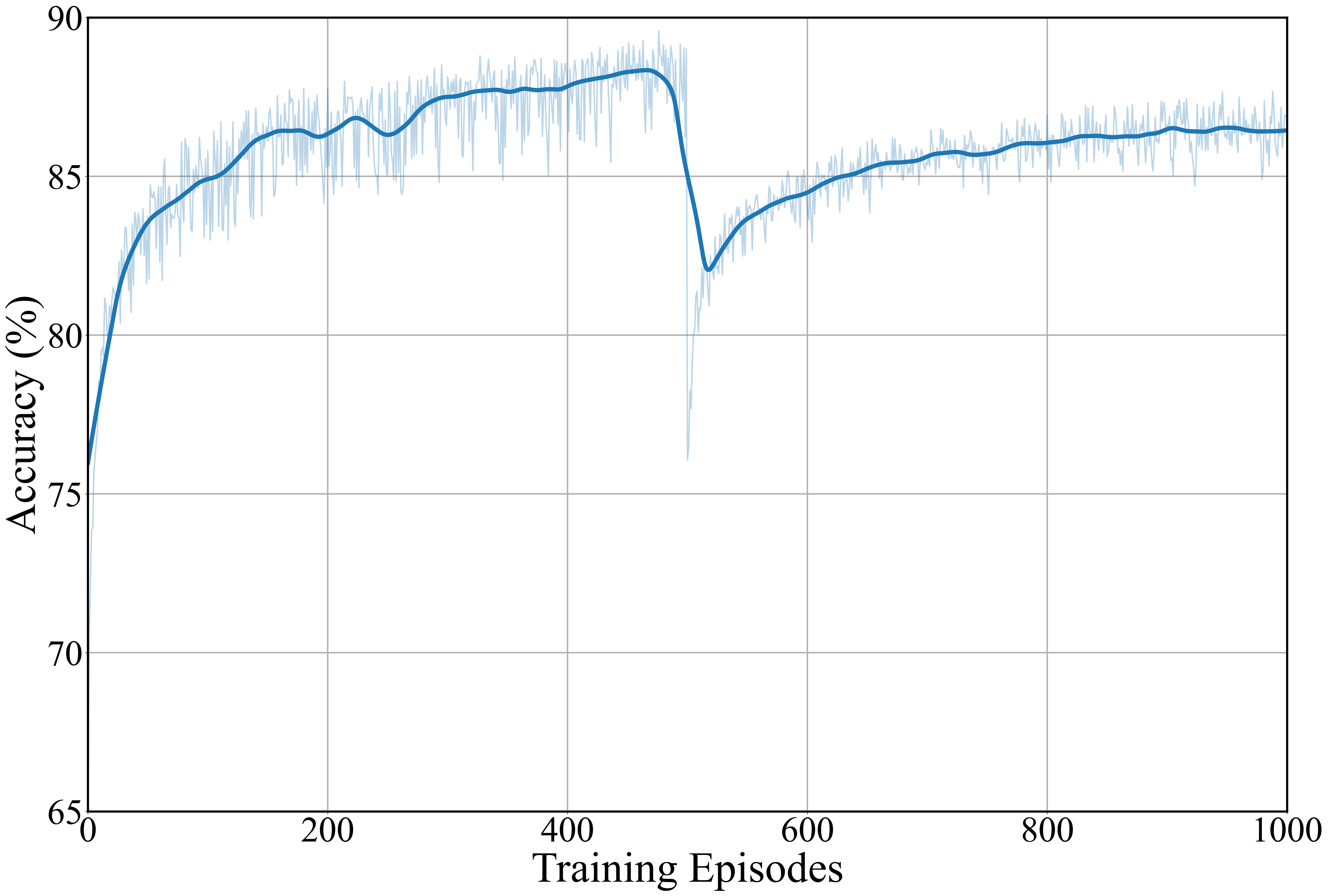}
    \caption{Validation accuracy during training process join a new client}
    \label{fig:new_client}
\end{figure}

In this experiment, we evaluate the capability of the proposed federated learning framework to adapt efficiently to new clients introduced during the training process and to resume learning with minimal disruption. This is a realistic requirement for long-running deployments of FL framework. The framework should quickly adapt to the new client without restarting from scratch and while maintaining stability and privacy. The experiment begins with four clients and training proceeds until convergence, after which a fifth client containing a previously unseen class is introduced. then continue training under the same protocol. Training then continues under the same protocol, and the raw accuracy trajectories and their LOWESS-smoothed curves are plotted in Fig. \ref{fig:new_client}.

From the result, the validation curve rises quickly and converges near 88\%, reflecting the efficiency of transfer learning in our design, shortening optimization and reducing communication. When the new client join the system at episode 500, the accuracy drops sharply to 76\%. The negative impact is due to abrupt label-distribution shift and temporary gradient misalignment introduced by a class that was unseen during prior rounds. However, Benefit by the dedicated techniques implemented in our proposed framework for accepting new client.  The accuracy rebounds to 85\% after 100 episodes, indicating fast adaptation without catastrophic forgetting. As a result, the system converges to 86\% as the new stable performance, slightly lower than the previous final accuracy before it joins. The result is consistent with previous test result in Secion. \ref{subsubsection:client_num}, that larger client numbers negatively affect the final performance because of more heterogeneous dataset distribution under non-IID.

The result in this test demonstrate that our framework is operationally robust to new client. it admits new clients in the middle of training, recovers quickly from the initial accuracy drop, and continues training to adapt to new optimal performance. This is essential for a long-running, real-world agricultural FL deployments where participants may join over time and the system must remain both effective and flexible.

%% file: Comparison_Table.tex
\begin{tabular}{cccccc}
\toprule
          \textbf{Class Name} & \textbf{Raw CLIP} & \textbf{FedTPG} & \textbf{Fed-CLIP} & \textbf{Proposed} & \textbf{Centralized Training}\\
\midrule
     asian-flatsedge &    21.50 &   1.41 &    40.31 &          95.93 &                                99.06 \\
   asiatic-dayflower &     2.33 &   0.17 &     6.83 &          97.18 &                                99.17 \\
                bean &     0.31 &  12.46 &     0.00 &          75.88 &                                93.45 \\
    bloodscale-sedge &     2.62 &   5.10 &    89.08 &          95.06 &                                98.41 \\
      cockspur-grass &     0.61 &   1.51 &     0.00 &          66.99 &                                95.27 \\
         cooper-leaf &     0.60 &   4.86 &     0.00 &          84.04 &                                96.97 \\
                corn &    31.70 &  58.40 &    55.90 &          87.44 &                                86.30 \\
early-barnyard-grass &     0.00 &  66.67 &    80.00 &          76.19 &                                42.86 \\
        fall-panicum &     4.16 &  31.94 &    92.85 &          94.11 &                                91.75 \\
        finger-grass &     0.11 &  19.89 &    91.97 &          92.62 &                                94.57 \\
      foxtail-millet &     7.19 &  11.82 &    89.14 &          75.86 &                                80.82 \\
           goosefoot &     0.00 &   3.76 &    84.93 &          94.11 &                                98.20 \\
        great-millet &     0.00 &   5.18 &    86.94 &          70.86 &                                89.35 \\
       green-foxtail &     0.00 &  13.44 &    66.49 &          83.66 &                                90.25 \\
          green-gram &     0.95 &   0.55 &    87.98 &          78.35 &                                93.21 \\
              henbit &     2.55 &  14.59 &    92.09 &          89.92 &                                96.15 \\
   indian-goosegrass &     0.50 &  19.03 &    53.15 &          81.77 &                                93.76 \\
         korean-dock &     0.00 &  27.62 &   100.00 &          99.34 &                                98.34 \\
       livid-pigweed &     4.33 &   0.80 &     0.00 &          79.97 &                                92.04 \\
    nipponicus-sedge &     0.00 &  34.38 &     5.48 &          99.45 &                                98.71 \\
              peanut &     0.42 &  54.74 &     0.00 &          94.84 &                                95.70 \\
             perilla &     0.00 &  56.89 &     0.00 &          86.85 &                                93.82 \\
           poa-annua &    12.50 &  33.33 &    12.50 &          85.71 &                                95.24 \\
        proso-millet &     8.64 &  30.61 &     0.00 &          76.32 &                                88.96 \\
            purslane &     8.18 &  33.98 &    69.89 &          95.05 &                                98.06 \\
            red-bean &     0.00 &  10.39 &    68.47 &          85.76 &                                90.09 \\
     redroot-pigweed &    12.14 &  35.54 &    65.93 &          80.77 &                                89.26 \\
              sesame &     0.00 &  32.94 &    69.54 &          75.98 &                                82.48 \\
      smooth-pigweed &    37.58 &  21.09 &    75.03 &          89.39 &                                94.30 \\
     white-goosefoot &     0.58 &  50.58 &    61.33 &          95.19 &                                96.65 \\
\midrule
    \textbf{Average Accuracy} &     \textbf{5.83} &  \textbf{20.26} &    \textbf{52.76} &          \textbf{86.62} &                                \textbf{93.55} \\
\bottomrule
\end{tabular}

%% file: Sections/conclusion.tex
\section{Conclusion}
\label{sec:conclusion}

In this paper, we introduce a novel federated learning framework called FedReplay, a federated learning framework that integrates a pretrained vision–language model with transfer learning for efficient and privacy-preserving agricultural image classification. By freezing the CLIP encoder and training only a lightweight Transformer-based classifier, the framework reduces communication overhead by about 98\% while retaining strong visual representation capability. The proposed feature replay, knowledge-distillation tuning, and row-gated aggregation strategies effectively address non-IID data challenges and enable seamless integration of late-joining clients.

Experimental results demonstrate that FedReplay achieves 86.6\% accuracy, exceeding baseline methods by more than fourfold while maintaining stability and scalability under heterogeneous agricultural data. These results highlight the framework’s potential as a practical foundation for intelligent and privacy-aware federated learning in smart agriculture applications.